\def\year{2021}\relax
\documentclass[letterpaper]{article} 
\usepackage{aaai20}  
\usepackage{times}  
\usepackage{helvet} 
\usepackage{courier}  
\usepackage[hyphens]{url}  
\usepackage{graphicx} 
\usepackage{natbib}  
\usepackage{caption} 
\usepackage{graphicx}
\usepackage{amsfonts}
\usepackage{multirow}
\usepackage{multicol}
\usepackage{url}
\usepackage{upgreek}
\usepackage{amssymb}
\usepackage[ruled,vlined]{algorithm2e}
\newcommand{\fixme}[1]{}

\title{Keyphrase Extraction with Dynamic Graph Convolutional Networks and Diversified Inference}
\author{
Haoyu Zhang\textsuperscript{\rm 1\thanks{\ \ Indicates equal contribution.}},
    Dingkun Long\textsuperscript{\rm 2*},
    Guangwei Xu\textsuperscript{\rm 2},
    Pengjun Xie\textsuperscript{\rm 2},
    Fei Huang\textsuperscript{\rm 2},
    Ji Wang\textsuperscript{\rm 1\thanks{\ \ Corresponding author.}}\\
    \textsuperscript{\rm 1}{State Key Laboratory of High Performance Computing, National University of Defense Technology}\\
    \textsuperscript{\rm 2}Alibaba Group\\
    \{zhanghaoyu10, wj\}@nudt.edu.cn, \{dingkun.ldk, kunka.xgw, chengchen.xpj, f.huang\}@alibaba-inc.com\\
}

\begin{document}

\maketitle

\begin{abstract}
Keyphrase extraction (KE) aims to summarize a set of phrases that accurately express a concept or a topic covered in a given document. Recently, Sequence-to-Sequence ({\it Seq2Seq}) based generative framework is widely used in KE task, and it has obtained competitive performance on various benchmarks. The main challenges of {\it Seq2Seq} methods lie in acquiring informative latent document representation and better modeling the compositionality of the target keyphrases set, which will directly affect the quality of generated keyphrases. In this paper, we propose to adopt the Dynamic Graph Convolutional Networks (DGCN) to solve the above two problems simultaneously. Concretely, we explore to integrate dependency trees with GCN for latent representation learning. Moreover, the graph structure in our model is dynamically modified during the learning process according to the generated keyphrases. To this end, our approach is able to explicitly learn the relations within the keyphrases collection and guarantee the information interchange between encoder and decoder in both directions. Extensive experiments on various KE benchmark datasets demonstrate the effectiveness of our approach.
\end{abstract}

\section{Introduction}
A keyphrase is a multi-word text representing highly abstractive information in a long document~\cite{DBLP:conf/acl/HasanN14}. Keyphrase extraction (KE) is a task that aims to generate an appropriate keyphrase set for the given document, thus helping to identify salient contents and concepts from the document. Recently, the KE task has attracted much research interest since it serves as an important component of many downstream applications such as text summarization~\cite{liu2009unsupervised}, document  classification~\cite{DBLP:conf/acl/HulthM06}, information retrieval~\cite{DBLP:conf/ijcnlp/KimKCOPS13} and question generation~\cite{DBLP:journals/corr/SubramanianWYT17}.

Early KE systems commonly operate in an extractive manner~\cite{DBLP:conf/emnlp/MihalceaT04,DBLP:conf/emnlp/MedelyanFW09}, which usually consists of two steps: 1) selecting candidates from the source document using heuristic rules,  and 2) ranking the candidates list to determine which is correct. However, the two-step ranking approaches are usually based on feature engineering, which is labor-intensive. Motivated by the progress in sequence-to-sequence applications of neural networks, KE research's focus has gradually shifted to deep learning methods. \citet{DBLP:conf/acl/MengZHHBC17} first formulate KE as a sequence generation problem and introduce an attentive Seq2Seq framework to generate the keyphrase sequence conditioned on the input document. Compared with traditional methods, the Seq2Seq based method achieves superior performance.

Seq2Seq based KE is exposed to two major challenges: 1) Document-level representation learning. For any Seq2Seq generative framework, the latent hidden representation is a very important factor, and its quality will directly affect the decoder's performance. In KE task, the input is commonly a long document instead of a sentence, which poses a greater challenge to latent representation learning. 2) Modeling the compositionality of keyphrases set. The elements in the keyphrase set are dependent and correlated. That is, better modeling the inherent composition embodied in the keyphrase set during the learning process will effectively boost the diversity and quality of final results. 

Recently, various approaches have been proposed to optimize the Seq2Seq generation framework in KE task. To learn a better latent representation, previous studies try to introduce different encoding structures (e.g., BiLSTM~\cite{DBLP:conf/acl/MengZHHBC17}, Graph Convolutional Networks~\cite{DBLP:conf/sigir/SunTDDN19} or using title guided encoder~\cite{DBLP:conf/aaai/ChenGZKL19}. Another line of work tries to learn the composition of keyphrase set by generating the concatenation of all target keyphrases~\cite{yuan2018one} or using coverage attention to reduce word repetition~\cite{DBLP:conf/acl/ZhaoZ19}. Existing approaches for KE mainly focus on one of the certain challenges as discussed. However, the two factors mentioned above are correlated.

The observations mentioned above motivate us to study a more general approach to the Seq2Seq based KE task. In this work, we propose to use a dynamic graph convolutional network (DGCN) to address the two issues above simultaneously. We explore to incorporate the dependency tree for document representation learning in the encoder part. The syntactic dependency tree can help to locate key information in a document. In practice, the document graph $G$ is constructed depending on the syntactic dependency tree, and then a convolution process will be operated over $G$.

On the other hand, we rethink the implication of compositionality in the keyphrase set. In the training process of generative models, whether a candidate keyphrase should be generated not only hinges on the document itself, but also depends on the keyphrases that have already been generated. Therefore, a dynamic graph updating mechanism is introduced to explicitly modeling the inter-dependency among keyphrases. In our method, the graph structure in the encoder part will be dynamically updated according to the keyphrases generated in the decoder part. Concretely, after one keyphrase is decoded, its information will be transferred to modify the edge weights in the document graph through a score function, and the latent hidden representation will also be updated. In this approach, we could dynamically ensure the information exchange between encoder and decoder parts in both directions.

The contribution of this work is three-fold: 
1) A novel generative framework, Div-DGCN, is proposed that leverages both the dynamic syntactic graph encoder and diversified inference process for KE.
2) A dynamic computation mechanism is adopted to model the compositionality in keyphrase set explicitly and then enhancing the information interchange between the encoder and decoder parts in the Seq2Seq architecture. 
3) Extensive experiments conducted on five benchmarks show that our proposed method is effective against competitive baselines on several metrics.

\section{Related Work}
Keyphrase extraction problem is usually carried out via extractive or generative methods. Conventional extractive methods usually use the two-step strategy that first extracts the candidate phrases using rules (hand-crafted or syntactic pattern matching) and then ranks them based on supervised or unsupervised methods~\cite{DBLP:conf/emnlp/MihalceaT04,DBLP:conf/emnlp/MedelyanFW09}.
\citet{DBLP:conf/aaai/GollapalliLY17} used sequence labeling models to extract keyphrases from the document.

Our model is based on Seq2Seq generative architecture. In this line of work, CopyRNN~\citet{DBLP:conf/acl/MengZHHBC17} is the first to cooperate copy mechanism to generate keyphrases. Since then, Seq2Seq based generative models have gradually become the mainstream in the KE task. \citet{DBLP:conf/aaai/ChenGZKL19} proposed a title-guided Seq2Seq network to enhance the latent document representation. \citet{DBLP:conf/acl/ZhaoZ19} introduced linguistic annotations for representation learning. Deep graph-based methods have been used for text representation learning in many NLP tasks such as text summarization~\cite{yasunaga2017graph}, semantic role labeling~\cite{marcheggiani2017encoding} and machine translation~\cite{bastings2017graph}. In KE task, ~\citet{DBLP:conf/sigir/SunTDDN19} proposed to leverage graph-based encoder to model document-level word salience globally. Compared to previous works, our model explores the syntactic structure and lets the global decoder side information flow to impact the encoder representation, thus generates more global-aware document latent representations. The idea to model dynamic node embedding was also studied in other tasks~\cite{DBLP:journals/corr/abs-1902-10191}, however, these methods evolve the model parameters instead of the graph structure, thus is different from ours.

There are also many studies focusing on the diversity of generated keywords. The catSeqD is an extension of catSeq with orthogonal regularization~\cite{DBLP:conf/nips/BousmalisTSKE16} and target encoding. \citet{DBLP:conf/emnlp/ChenZ0YL18} further proposed a review mechanism to model the correlation between the keyphrases explicitly. Lately, ~\citet{DBLP:conf/acl/ChanCWK19} proposed a reinforcement learning based fine-tuning method, which fine-tunes the pre-trained models with adaptive rewards for generating more sufficient and accurate keyphrases. \citet{DBLP:journals/corr/abs-2004-08511} designed a soft/hard exclusion mechanism to enhance the diversity. 
The idea of enhancing diversity is roughly selecting results that have not been generated or ensuring the whole results covering the main semantic contents. 

\section{Problem Definition}
Keyphrases are usually divided into two categories: \textit{present} and \textit{absent}, determined by whether the phrase appears in the source document~\cite{DBLP:conf/acl/MengZHHBC17, DBLP:conf/acl/ChanCWK19}. In this work, we concentrate on present keyphrase generation problem (keyphrase extraction), which can be formulated as follows: given source document $\mathcal{D}$ = $\{x_1, x_2, \cdots, x_{l}\}$ with $l$ words, the ground-truth keyphrase set is $\mathcal{Y}$ = $\{y_1, y_2, \cdots, y_n\}$ with $n$ keyphrases. We split the keyphrase set into present keyphrases $\mathcal{Y}^p$ and absent keyphrases $\mathcal{Y}^a$. The target of present keyphrase generation is $\mathcal{Y}^p$, and we denote $\mathcal{Y}^p$ as $\mathcal{Y}$ in the rest of this paper for brevity. Therefore, each keyphrase $y_j$ in $\mathcal{Y}$ is sequence of words: $y_j = \{x^j_{m_1}, 
x^j_{2}, \cdots, x^j_{m_j}\}$. Since we formulate the problem as a sequence generation problem, all present keyphrases are concatenated by a special token ``\mbox{SEP}'' to compose the target sequence with another token ``\mbox{EOS}'' in the end: $\{y_1,\mbox{SEP},y_2,\mbox{SEP},\cdots,\mbox{EOS}\}$.

\section{Method}
\begin{figure*}[t]
\centering
\includegraphics[width=0.95\textwidth]{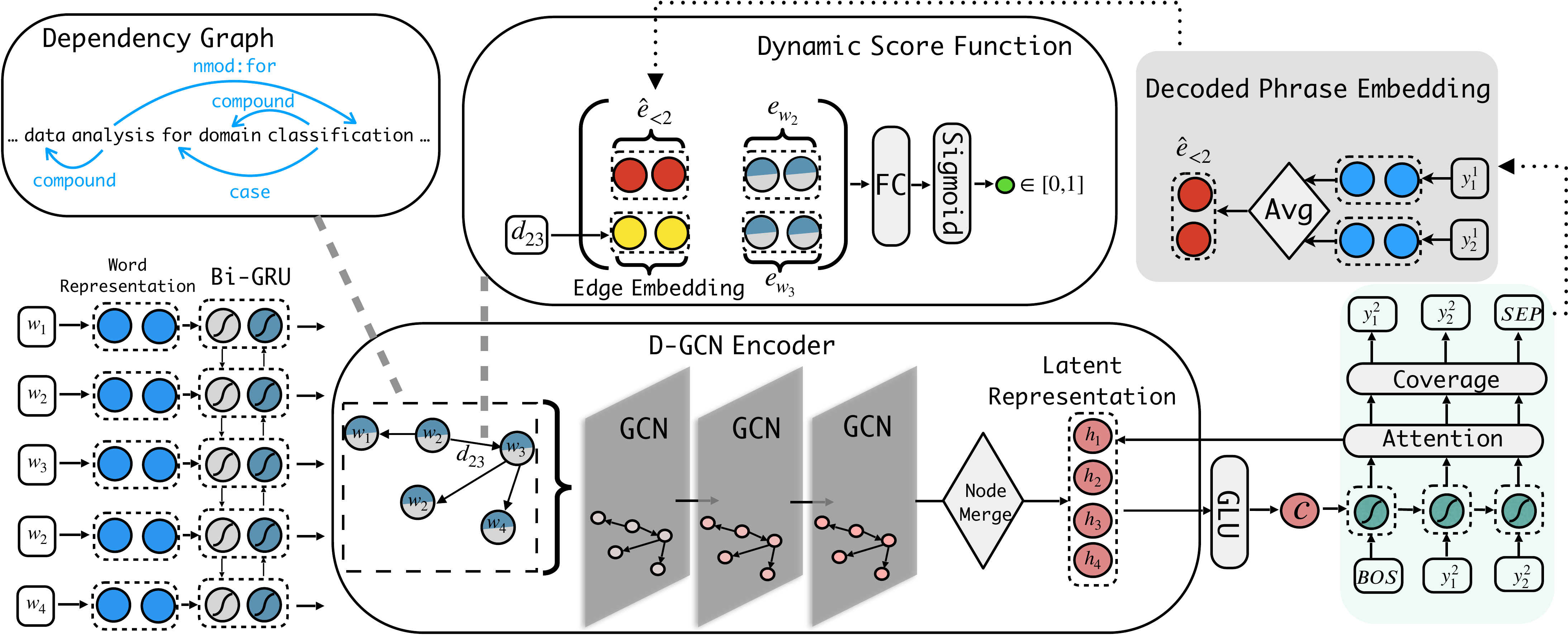}
\caption{An overall architecture of our proposed Div-DGCN model. In this example, $\mathcal{D}$ = $\{w_1, w_2, w_3, w_2, w_4\}$ and the first keyphrase $\{y^1_1, y^1_2\}$ has been generated.}
\label{fig:model}
\end{figure*}

Based on the Seq2Seq framework, the overall architecture of the proposed Div-DGCN framework is depicted in Figure~\ref{fig:model}. It mainly consists of a syntactic GCN encoder, a GRU based decoder, and a dynamic computation mechanism that bidirectionally associates the encoder and decoder.

\subsection{Syntactic GCN Encoder}
As analyzed in~\cite{Meng2020AnES}, keyword extraction model based on  Seq2Seq architecture tends to generate repetitive candidate sequences, many of which are parent-child strings, and which will obviously hurt the diversity of generated results. To address this problem, we introduce syntactic information to help the encoder better locate semantically completed candidate phrases in sentences (more details can be found in the Appendix). 

The syntactic GCN encoder aims at encoding a variable-length document $\mathcal{D}$ into a latent representations $H = \{h_1, h_2, \cdots, h_{l}\}, h_j \in \mathbb{R}^{d_h}$, where $d_h$ is the latent vector dimension. We first map each word $x_i$ in the source document into word embedding $e^w_i\in \mathbb{R}^{d_w}$, POS embedding $e^{pos}_i \in \mathbb{R}^{d_{pos}}$ (using its POS annotation) and position embedding~\cite{DBLP:conf/nips/VaswaniSPUJGKP17} $e^p_i\in \mathbb{R}^{d_p}$. The final word representation $e_i$ is the concatenation of the three parts $[e^w_i; e^{pos}_i; e^p_i]$, $d_w, d_{pos}$ and $d_p$ represent the embedding size of each part respectively. Before the GCN module, we apply a bidirectional GRU \cite{chung2014empirical} (BiGRU) to access the context features. Using the word sequence representation as input, the BiGRU network generates context-aware hidden states in both forward and backward directions. We can concatenate the corresponding representations in two directions as the output representations $H^0 = \{h^0_1, h^0_2, \cdots, h^0_l\}$. 

\noindent \textbf{Syntactic Graph Construction} Given a source document $\mathcal{D}$ with $l$ words, we first split the document into sentences and get the sentence level syntactic dependency tree using Stanford Dependency Parser~\cite{DBLP:journals/corr/abs-2003-07082}. The dependency trees can be interpreted as word-level directed graph, where each node represents one word in the source document, and each edge represents a syntactic dependency path between two words. Mathematically, the directed graph $G$ of the entire document can be represented as a sparse adjacency matrix $A \in \mathbb{R}^{l\times l}$. Previous works using syntactic trees and graph convolutional networks usually treated matrix $A$ as a binary value matrix~\cite{marcheggiani-titov-2017-encoding, sun-etal-2019-aspect}. Namely, $A_{ij} = 1$ if there exists a dependency path from word $i$ to $j$ or from word $j$ to $i$, and otherwise $A_{ij} = 0$. In this work, to fully leverage the dependency tree structure information to represent the association between words pair, we use continuous value to indicate the closeness of such association. In particular, these continuous values are calculated by a score function that incorporating both the word embeddings and the dependency type. For the edge between word $i$ and word $j$ with dependency type $d_{ij}$, the weight $A_{ij}$ in matrix $A$ is calculated by:
\begin{equation}
    A_{ij} = \sigma(W_e([e_i; e_j; e^t_{d_{ij}}])) \label{eq:graph_construct}
\end{equation}
where $e^t_{d_{ij}}\in \mathbb{R}^{d_t}$ is the dependency type embedding, $W_e$ is a parameter matrix and $\sigma$ represents the sigmoid function. All calculated weights will located between 0 and 1. If there is no dependency path between word $i$ and $j$, $A_{ij} = 0$. 

\noindent \textbf{Graph Convolutional Networks} After constructing the syntactic graphs $G$, we employ a multi-layer Graph Convolutional Network (GCN) to get the document representation, as shown in Figure~\ref{fig:model}. The output of the BiGRU network $H^0$ will be used to initial node embeddings in $G$. In each layer, the GCN encoder will only consider each node's one-hop neighborhood, then aggregate the neighbors' information when updating its representation. The update process in the $k$-th layer GCN can be represented as:
\begin{equation}
h^k_i = {\rm ReLU} (\sum_{j=1}^l \frac{1}{d_i} A_{ij} (W^k h_j^{k-1} + b^k)) \label{eq:gcn}
\end{equation}
where $h_i^k$ is the $i$-th node representation at the $k$-th layer, $W^k, b^k$ are layer-specific trainable weight/bias parameters, $d_i$ is a normalization constant denoting the degree of node $i$, and we use ReLU~\cite{nair2010rectified} as the activation function. For a $N$-layer GCN, the node representations are $H^N = \{h^N_1, \cdots, h^N_l\}$. In the rest of the paper, we denote $H^N$ as $H$.

For the keyphrase generation task, we seek to acquire a continuous representation of the entire document. However, the dependency trees are in sentence-level. Thus, we add a node merge layer to enhance the information interchange between sentences. Specifically, words share the same stem word are aggregated into only one representation through an average pooling function. Thus, the new sequential representation is $H = \{h_1, \cdots, h_{l'}\}$, where $l'$ notes the sequential length after the merge operation. 

The final document representation (or the entire context vector) $c$ is computed by a residual Gated Linear Unit (GLU) layer~\cite{DBLP:conf/cvpr/HeZRS16} and an averaging layer:
\begin{eqnarray}
    H & = &  H + (W_kH \otimes \sigma (W_lH)) \\
     c &=& \frac{1}{|l'|}\sum_{i=1} H_i
\end{eqnarray}
where $\otimes$ denotes element-wise multiplication, and $c$ is used to initialize the decoder hidden state.

\subsection{Copy Decoder}
The body of the decoder is a GRU network. At time step $t$, the hidden state $s_t$ in GRU is updated based on the embedding of previous decoded word $e^w_{y_{t-1}}$ and the hidden state at time step $t-1$: $s_t = f(s_{t-1}, e^w_{y_{t-1}}), s_0 = c$. To cope with the Out-of-Vocabulary (OOV) challenge, we adopt a Pointer Generator~\cite{DBLP:conf/acl/SeeLM17} style copy mechanism. For each document $\mathcal{D}$, a dynamic vocabulary $\mathcal{V}_d$ is calculated to cover all tokens in $\mathcal{D}$ as well as the 
pre-build static vocabulary $\mathcal{V}$. By attending the hidden state $s_t$ to the latent document representations $H$, the copy distribution over $\mathcal{V}_d$ is calculated as the final predictive distribution. In addition, we observe that keyphrases are located in different parts of the source document. Therefore, we adopt a coverage mechanism~\cite{DBLP:conf/emnlp/ChenZ0YL18} to prevent the predictive distribution from locating on a small portion of $\mathcal{D}$. 

We utilize the beam search to decode keyphrase sequences. Beam search stores the top-B highly scoring candidates at each time step, where B is known as the beam width. When decoding the $p$-th keyphrase at its $t$-th word, the set of B candidate sequences can be denoted as $Y^p_{t} = \{\textbf{y}^j_{1, [t]}, \cdots, \textbf{y}^p_{B, [t]}\}, \textbf{y}^p_{b, [t]}\in \mathbb{R}^{t}$, thus, all choices at $t$-th time step over the dynamic vocabulary are $\mathcal{Y}^p_{t} = Y^p_{t-1}\times \mathcal{V}_d$. To reduce the search space, we leverage a phrase-level beam search method. The phrase-level beam search decodes each keyphrase individually, and once a keyphrase is decoded, the candidate sequence with the highest score will be picked. In this way, the optimization objective is represented as: 
\begin{equation}
Y^p_{t} = \mathop{\arg\max}_{\textbf{y}_{1, [t]}, \cdots, \textbf{y}_{B, [t]}\in \mathcal{Y}_t} \sum_{b\in [B]} \Theta (\textbf{y}^p_{b, [t]}) \label{eq:beam_score}
\end{equation}
which indicates that we aim to select $B$ sequences with maximum scores from $B \cdot |\mathcal{V}_d|$ members in $\mathcal{Y}^p_t$. For the sequence score function $\Theta(\cdot)$, we use the negative log probability sum with length penalty.

\noindent \textbf{Diversified Inference} To enhance the semantic diversity of generated keyphrases. Motivated by previous works~\cite{DBLP:journals/corr/LiMJ16,DBLP:conf/ijcai/ShiCQH18}, we propose a diversified inference algorithm for the phrase-level beam search via adding two terms. First, we add a phrase-level dissimilarity penalty (DP) term $\mbox{DP}(\textbf{y}^p_{b, [t]}, \textbf{y}^{<p})$ to the score function $\Theta$,  where $\textbf{y}^{<p}$ denotes the first decoded $p-1$ keyphrases. A key motivation behind the dissimilarity term is that the current keyphrase should be different from previous decoded ones. Hence, we use the uni-gram and bi-gram overlap ratio to measure the dissimilarity between two sequences. On the other hand, as the standard beam search selects sequence based on the accumulated log probability of each token, nodes with very high score might dominate the search process, which will degrade the candidate search space and generate similar keyphrases. Thus, we introduce a sibling penalty (SP) term $\mbox{SP}(y^p_{b, t})$ whose value is the log-probability rank of the $t$-th candidate token in the candidate sequence over $|\mathcal{V}_d|$. By incorporating the sibling penalty, we can filter the low-score candidates that have high-score ancestors. To this end, the final sequence score function is: 
\begin{equation}
    \hat \Theta (\textbf{y}^p_{b, [t]}, \textbf{y}^{<p}) = \Theta (\cdot) + \lambda_1 \mbox{DP}(\cdot) - \lambda_2 \mbox{SP}(y^p_{b, t}) \label{eq:bs_func}
\end{equation}
where $\lambda_1$ and $\lambda_2$ are hyper-parameters, which control the importance of each term. The diversified inference process of a single keyphrase is illustrated in Algorithm~\ref{algo:1}.

\begin{algorithm}[t]
\caption{Inference Procedure}
\label{algo:1}
\small
 Input:\\
 - $\mathcal{D} = \{x_1, \cdots, x_l\}$ ; $\textbf{y}^{<p} = \{y^1, \cdots, y^{p-1}\}$ \\
 - Model parameters: $\theta$ \\
 
 Perform beam search for the $p$-th keyphrase using beam width $B$ \\
 $H^p, c^p = \mbox{ENCODER}(D, \textbf{y}^{<p}; \theta)$ \\
\For{$t=1,\ \ldots \,T$}{
    $P(y^p_{b, t}) = \mbox{DECODER}(y^p_{b, <t}, H^p, c^p; \theta)$ \\
    $\mathcal{Y}^p_{t} =  \mathcal{Y}^p_{t-1} \times P(y^p_{b, t})$ \\
    $\hat \Theta(\textbf{y}_{b,[t]}^p) \leftarrow \Theta(\textbf{y}_{b,[t]}^p) + \lambda_1 \mbox{DP}(\textbf{y}_{b,[t]}^p, \textbf{y}^{<p}) - \lambda_2 \mbox{SP}(y^p_{b, t})$  \\ 
        $Y_{t}^p \leftarrow \mathop{\arg\max}_{(\textbf{y}_{1,[t]}^p, \dots, \textbf{y}_{B,[t]}^p)} \sum_{b\in[B]}\hat \Theta(\textbf{y}_{b,[t]}^p)$ \\
}
Return $y^p = \mathop{\arg\max} (Y^p_{T})$ using $\hat \Theta(\cdot)$ as the predicted $p$-th keyphrase.
\end{algorithm}

\subsection{Dynamic Graph Convolutional Networks}
The use of syntactic tree information and graph convolutional networks allows us to obtain a better document representation. However, to generate high-quality keyphrases, it is also important to model the inter-relationship between keyphrases. To address this issue, previous attempts mainly focus on modifying the decoder structure. The modification of decoder part means that the decoder process still depends on the static hidden representation $c$. In this work, we model the inter-relationship between keyphrases set in a global view. As discussed, the decoding process of $p$-th keyphrase depends both on the source document and the previously decoded keyphrases. This observation motivates us to transform the hidden representation to evolve together with the decoding process dynamically. Moreover, the score function design in graph $G$ allows us to carry out this dynamic process at a very low cost. 

Specifically, for the input graph $G$, we modify the edge weights in matrix $A$ after decoding each keyphrase. The score function in Eq.~\ref{eq:graph_construct} is extended to the following form:
\begin{equation}
    A^p_{ij} = \sigma (W_e([ e_i; e_j; e^t_{d_{ij}}; \hat e_{<p} ] ) ) \label{eq:d_score_function}
\end{equation}
when decoding the $p$-th keyphrase, the decoded keyphrases word sequence is $y^{<p} = \{y^1_{1}, \cdots, y^1_{l_1}, \cdots, y^{p-1}_{1}, \cdots, y^{p-1}_{l_{p-1}}\}$, and the decoded phrase embedding $\hat e_{<p}$ in Eq.~\ref{eq:d_score_function} represents the average pooling results of all decoded word embeddings. When $p$=$1$, $\hat e_{<p}$ is a zero vector. That is, the relatedness between two words are not only determined by its word embedding and dependency types corresponds, but also depends on the previously decoded contents. Once a keyphrase is generated, we will updated the 
adjacent matrix $A$ of $G$ through Eq.~\ref{eq:d_score_function}. In this way, the hidden representation $c$ will also be updated. As a result, during the encoding process, the graph $G$ is capable to dynamically extract information which is informative for the decoder part. 

The dynamic computation mechanism enable us to transmit the information of decoder to the encoder, then informs the ungenerated keyphrases of what have been already generated. Consequently, we explicitly model the relationship between each element in the keyphrases set.

\subsection{Training}
Given training data $\{\mathcal{D},\mathcal{Y}\}$, $\mathcal{D}$ notes the document and $\mathcal{Y}$ represents the concatenated keyphrases. The loss function with parameters $\theta$ is the average negative log-likelihood on all ground-truth words in the keyphrase sequences:
\begin{equation}
  \mathcal{L}(\theta) = -\sum_{p=1}^{n}\sum_{j=1}^{l_p}  log(P(y^p_j|y^p_{<j}, y^{<p}, \mathcal{D};\theta))
\end{equation}
\section{Experiment}
We thoroughly evaluate the performance of our method on five benchmarks, and we use several competitive approaches as baselines. Also, several auxiliary experiments are conducted to analyze the effectiveness of our method. All experiments are repeated three times using different random seeds and the averaged results are reported.

\subsection{Implementation Details} 

We utilize Stanford Parser~\cite{DBLP:journals/corr/abs-2003-07082} for the dependency parsing and POS annotation.
We use the Pre-trained 300 dimension FasText word embeddings~\cite{DBLP:journals/corr/BojanowskiGJM16} and keep the word embeddings learnable during training.
Our model variants are trained using the Adam optimizer~\cite{DBLP:journals/corr/KingmaB14} with a batch size 128 and an initial learning rate $0.001$.
During training, we use a dropout rate of $0.2$ and a gradient clipping threshold of $0.2$.
We train the model for $20$ epochs, and every $2000$ iterations the validation perplexity (ppl) are evaluated.
The learning rate is reduced by half if the validation ppl does not drop, three contiguous stagnant ppl will trigger the early stop of training.
All characters in the document and keyphrase set are lower-cased and all digits are replaced with a special token.

\noindent \textbf{Decoding Process}
We use exactly the same pre-process, post-process and performance evaluation processes with previous studies~\cite{yuan2018one,DBLP:conf/acl/ChanCWK19}.
Following~\cite{DBLP:conf/sigir/SunTDDN19}, we set beam search with beam  width $100$ in the decoding module on all test sets. If the termination symbol '[EOS]' is encountered or the number of keyphrases genrated exceeds the pre-defined maximum number, the decoding process will be terminated.
To ensure the fairness of the comparison, the testing process is consistent with~\cite{DBLP:conf/sigir/SunTDDN19} and all repeated keyphrases in the generated sequences are removed before evaluation.

\noindent \textbf{Hyper-parameters}
Table~\ref{table:hyper} in the Appendix shows all hyper-parameters used in our model variants. The values of these hyper-parameters are chosen empirically. In the beam search process, we choose the same length penalty factor with previous work~\cite{DBLP:conf/sigir/SunTDDN19}. The two diversified factors $\lambda_1$ and $\lambda_2$ are chosen separately depending on the performance improvement on the validation set. The detailed performance curves are also included in the Appendix.

\noindent \textbf{Datasets}
We conduct experiments on five scientific article datasets, including Kp20k~\cite{DBLP:conf/acl/MengZHHBC17}, Inspec~\cite{DBLP:conf/emnlp/Hulth03}, Krapivin~\cite{krapivin2009large}, NUS~\cite{DBLP:conf/icadl/NguyenK07} and SemEval~\cite{DBLP:conf/semeval/KimMKB10}. 
Each sample from these datasets consists of the title, abstract, and the target keyphrases. 
We concatenate the title and abstract to compose the input document.
Following previous studies, we train our model on the Kp20k dataset.
Then we evaluate our model on five benchmark test datasets.
Table~\ref{tab:statistics} illustrates more details of the datasets.

\begin{table}[t]
\centering
\scalebox{1.0}{
\begin{tabular}{|c|c|c|c|} 
\hline
\textbf{Dataset} & \textbf{\#samples} & \textbf{\#present} & \textbf{length}\\
\hline
\multicolumn{4}{|c|}{\textbf{training data}}\\
\hline
Kp20k & 464,676 & 2.94 & 2.01\\
\hline
\multicolumn{4}{|c|}{\textbf{validation data}}\\
\hline
Kp20k& 20,000 & 3.49 & 1.86 \\
\hline
\multicolumn{4}{|c|}{\textbf{test data}}\\
\hline
Inspec & 500 & 7.20 & 2.40\\
\hline
NUS& 211 & 5.64 & 1.93\\
\hline
SemEval& 100 & 6.12 & 2.07\\
\hline
Krapivin& 400 & 3.24 & 1.86\\
\hline
Kp20k& 20,000 & 3.31 & 1.86\\
\hline
\end{tabular}}
\caption{Statistics of five datasets.}
\label{tab:statistics}
\end{table}

\begin{table*}[t]
\centering
\scalebox{0.98}{
\begin{tabular}{l|p{0.8cm}p{0.8cm}| p{0.8cm}p{0.8cm} | p{0.8cm}p{0.8cm} | p{0.8cm}p{0.8cm} | p{0.8cm}p{0.8cm}}
\hline
\multicolumn{1}{c|}{\multirow{2}{*}{\textbf{Model}}} & \multicolumn{2}{c|}{\textbf{Inspec}} & \multicolumn{2}{c|}{\textbf{Krapivin}} & \multicolumn{2}{c|}{\textbf{NUS}} & \multicolumn{2}{c|}{\textbf{SemEval}} & \multicolumn{2}{c}{\textbf{KP20k}} \\
\multicolumn{1}{c|}{} & $F_{1}@M$   & $F_{1}@5$   & $F_{1}@M$    & $F_{1}@5$    & $F_{1}@M$  & $F_{1}@5$ & $F_{1}@M$    & $F_{1}@5$   & $F_{1}@M$   & $F_{1}@5$  \\
\hline 
Transformer~\cite{DBLP:conf/nips/VaswaniSPUJGKP17} &0.254 & 0.210 & 0.328 & 0.252 & - & - & 0.310 & 0.257 & 0.360 & 0.282 \\
catSeq~\cite{yuan2018one}      &0.262 & 0.225 & 0.354 & 0.269 & 0.397 & 0.323& 0.283 & 0.242 & 0.367 & 0.291 \\
catSeqD~\cite{yuan2018one}     &0.263 & 0.219 & 0.349 & 0.264 & 0.394 & 0.321& 0.274 & 0.233 & 0.363 & 0.285 \\
catSeqCorr~\cite{DBLP:conf/emnlp/ChenZ0YL18}   &0.269 & 0.227 & 0.349 & 0.265 & 0.390 & 0.319& 0.290 & 0.246 & 0.365 & 0.289 \\
catSeqTG~\cite{DBLP:conf/aaai/ChenGZKL19}    &0.270 & 0.229 & 0.366 & 0.282 & 0.393 & 0.325& 0.290 & 0.246 & 0.366 & 0.292 \\
catSeqD-$2RF_{1}$~\cite{DBLP:conf/acl/ChanCWK19}    &0.292 &0.242 & \textbf{0.360} & 0.282 &\textbf{0.419} & 0.353& 0.316 & 0.272& \textbf{0.379}& 0.305 \\
ExHiRD-h~\cite{DBLP:journals/corr/abs-2004-08511} &0.291 &0.253 & 0.347 & 0.286 & - & - & \textbf{0.335} & 0.284 & \textbf{0.374} & \textbf{0.311} \\
\hline
GCN   & 0.312 & 0.261 & 0.331 & 0.262 & 0.391 & 0.325& 0.303 & 0.264& 0.350 & 0.281 \\
Div-GCN   & 0.360 & 0.331 & 0.324 & 0.280 & 0.389 & 0.371& 0.317 & 0.289& 0.343 & 0.304 \\
DGCN   &0.327 &0.275 & 0.354 & 0.284 & 0.397 & 0.328& 0.319 & 0.280& 0.359 & 0.290 \\
Div-DGCN   &\textbf{0.376} &\textbf{0.348} & 0.345 & \textbf{0.313} & 0.391 & \textbf{0.379}& 0.323 & \textbf{0.320}& 0.349 & \textbf{0.313} \\
\hline
\hline
\multicolumn{11}{c}{\textbf{Normalized Discounted Cumulative Gain @10 (NDCG@10)}}\\
\hline
DivGraphPointer~\cite{DBLP:conf/sigir/SunTDDN19} & \multicolumn{2}{c|}{0.503} & \multicolumn{2}{|c|}{0.591} & \multicolumn{2}{|c|}{0.518} & \multicolumn{2}{|c|}{0.534} & \multicolumn{2}{|c}{0.532}\\
\hline
CatSeqTG-$2RF_{1}$~\cite{DBLP:conf/acl/ChanCWK19} & \multicolumn{2}{c|}{0.614} & \multicolumn{2}{|c|}{0.586} & \multicolumn{2}{|c|}{0.780} & \multicolumn{2}{|c|}{0.677} & \multicolumn{2}{|c}{0.592}\\
\hline
GCN & \multicolumn{2}{c|}{0.657} & \multicolumn{2}{|c|}{0.582} & \multicolumn{2}{|c|}{0.787} & \multicolumn{2}{|c|}{0.677} & \multicolumn{2}{|c}{0.590}\\
\hline
DGCN & \multicolumn{2}{c|}{0.663} & \multicolumn{2}{|c|}{0.597} & \multicolumn{2}{|c|}{\textbf{0.797}} & \multicolumn{2}{|c|}{\textbf{0.704}} & \multicolumn{2}{|c}{\textbf{0.616}}\\
\hline
Div-DGCN & \multicolumn{2}{c|}{\textbf{0.692}} & \multicolumn{2}{|c|}{\textbf{0.611}} & \multicolumn{2}{|c|}{\textbf{0.797}} & \multicolumn{2}{|c|}{0.701} & \multicolumn{2}{|c}{\textbf{0.620}}\\
\hline
\end{tabular}
} 
\caption{Results of present keyphrase prediction on five datasets. The best results are in bold. ``GCN'' denotes the model using syntactic GCN encoder. ``DGCN'' denotes the model using syntactic GCN encoder and dynamic computation mechanism. ``Div-DGCN'' and ``Div-GCN'' are the corresponding model variants decoding with diversified inference.}
\label{table:present-result}
\end{table*}

\noindent \textbf{Baselines}
We compare our methods with state-of-the-art KE systems include:
CatSeq / CatSeqD~\cite{yuan2018one}, CatSeqTG~\cite{DBLP:conf/aaai/ChenGZKL19}, CatSeqD-2RF${_1}$ / CatSeqTG-2RF${_1}$~\cite{DBLP:conf/acl/ChanCWK19}, DivGraphPointer~\cite{DBLP:conf/sigir/SunTDDN19} and ExHiRD-h~\cite{DBLP:journals/corr/abs-2004-08511}.
In fact, excluding the syntactic DGCN encoder and the diversified inference, our model is equal to CatSeqD except we only use copy probability.

\noindent \textbf{Evaluation Metrics}
We use $F_1$@$M$ proposed in~\cite{yuan2018one} and $F_1$@5 as our evaluation metrics. $F_1$@$M$ calculates the F1 score by comparing all the generated keyphrases with ground-truth keyphrase. Namely, $M$ is determined by the number of generated keyphrases. When calculating $F_1$@$5$, if the model predicts less than $5$ keyphrases, random wrong answers are appended for evaluation. Marco average is used to aggregate the evaluation scores for all samples. Note that we use Porter Stemmer for preprocessing to determine whether the two keyphrases are matching.
These two metrics are \textbf{commonly used} in previous works including~\cite{DBLP:conf/acl/ChanCWK19,DBLP:journals/corr/abs-2004-08511}.
We also introduce a ranking-based metric, Normalized Discounted Cumulative Gain (NDCG)~\cite{DBLP:conf/colt/WangWLHL13} to evaluate our model performances. Finally, we add experimental results using another version of F1@5 and F1@10 adopted by~\cite{DBLP:journals/corr/abs-1909-03590} as evaluation metrics and put it in the Appendix.

\begin{figure}[t]
\centering
\includegraphics[width=0.45\textwidth]{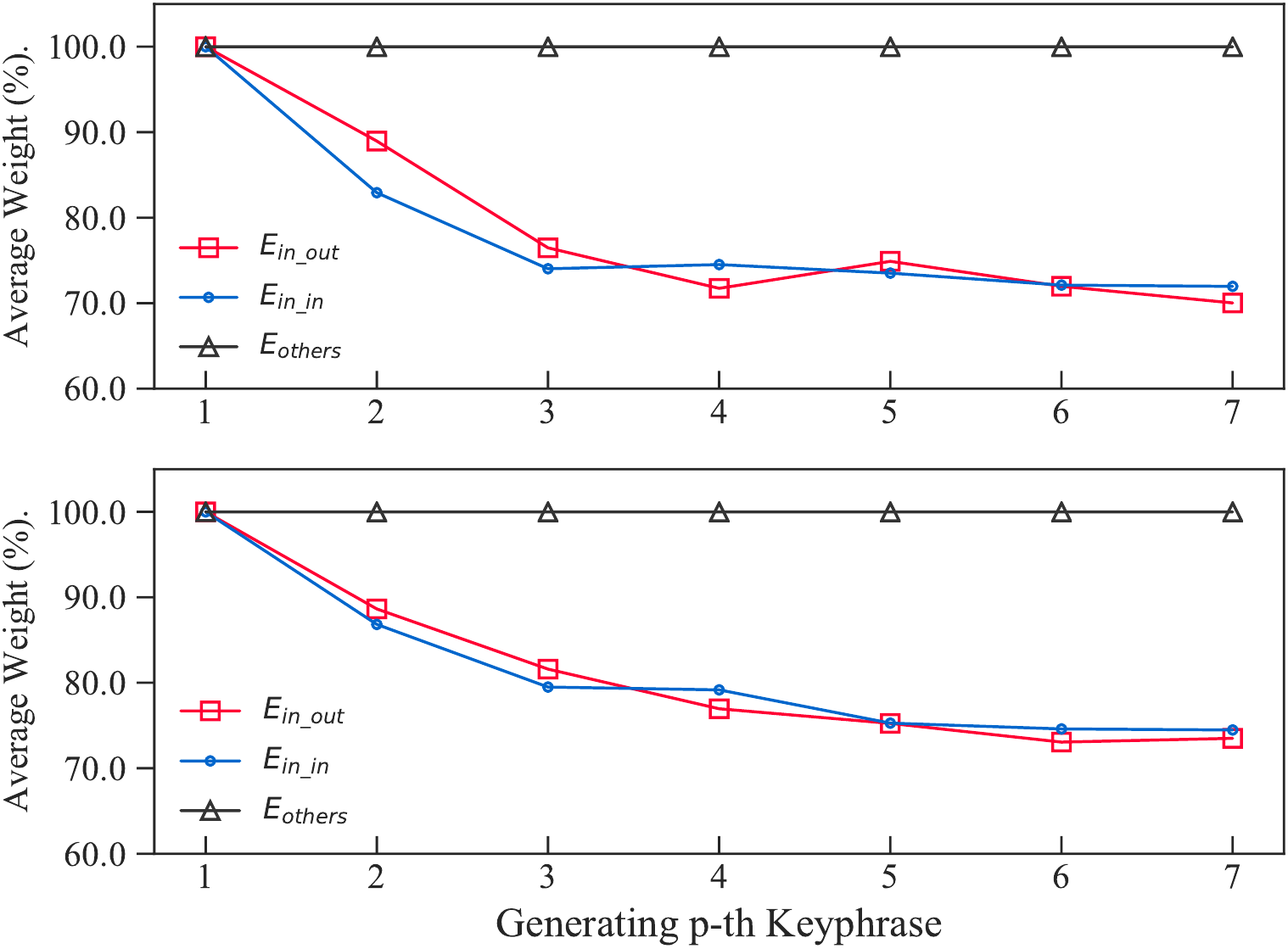}
\caption{Average weight trends of particular edge types on Inspec (upper) and NUS (lower) test set.}
\label{fig:graph_weight_trends}
\end{figure}
\begin{figure*}[t]
\centering
\includegraphics[width=0.98\textwidth]{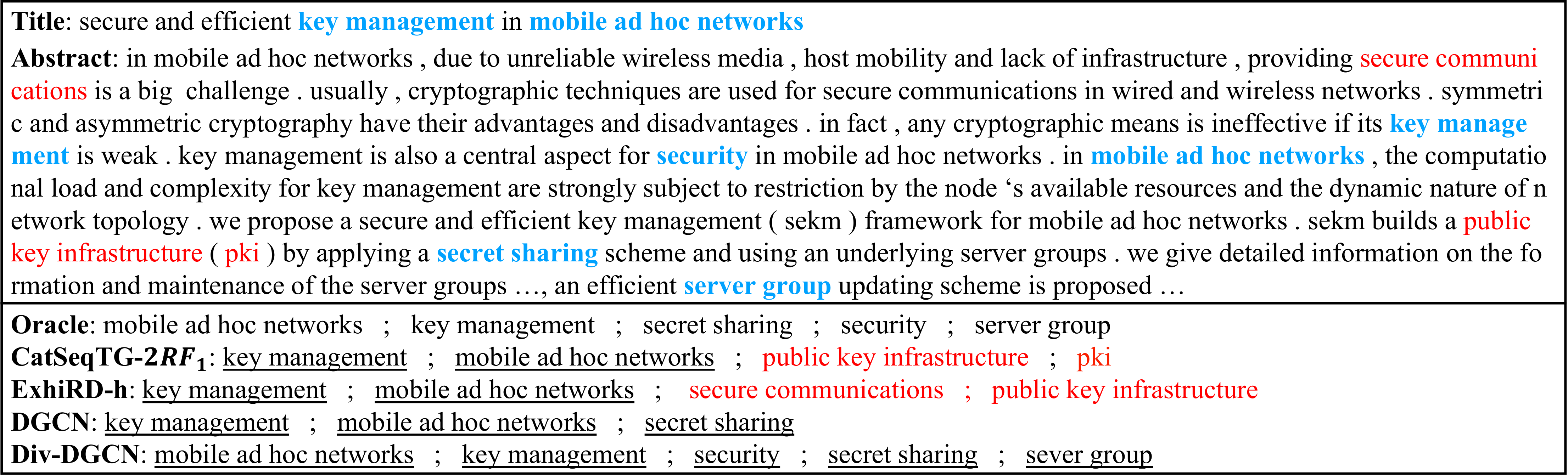}
\caption{A prediction example of baselines and our model variants. The correct keyphrases are underlined.}
\label{fig:case_study}
\end{figure*}

\begin{table}[t]
\centering
\scalebox{1.0}{
\begin{tabular}{l|c|c|c|c}
\hline \hline
Datasets & \# Nodes & \# Edges & \# E$_{in\_in}$ & \# E$_{in\_out}$ \\
\hline \hline
Inspec                                               & 145.27     &  374.07       &  40.64        &  75.82         \\
\hline
Krapivin                                              & 197.24    &  532.32       &  19.66        &  72.48         \\\hline
NUS                                              & 240.81     &  627.68       &  34.97        &  109.00         \\\hline
SemEval                                              & 245.72    & 679.97       &  43.98        &  131.78         \\
\hline
\end{tabular}
}
\caption{The node and edge statistics on four test sets.}
\label{table:graph-statistics}
\end{table}

\subsection{Overall Results}
The final results are reported in Table~\ref{table:present-result}, from which we can observe that: our Div-DGCN model shows competitive performances compared with previously proposed methods on all five datasets. Our model exceeds the SOTA method on $F_1@5$ performances by a large margin (average \textbf{13.5\%} performance gain on 5 test sets), which proves the diversity of our generated results. In terms of the $F_1$@$M$ metric, the performance of Div-DGCN is slightly inferior to the SOTA method, but there is still a significant improvement compared to the baseline model. Comparing with previous works using $F_1$@5 and $F_1$@10 (see Appendix) as evaluation metrics, we observe that our Div-DGCN model presents a promising improvement against previous methods.

In the KE task, the Seq2Seq architecture tends to generate a larger number of candidates, but few of them (less than $5\%$) are unique~\cite{Meng2020AnES}. These observations prove that the improvement on $F_1@5$ of Div-DGCN is due to the improvement of the moded\'s ability on generating accurate and diversify accurate keyphrases. On the other hand, our method focuses on generating keyphrases with higher diversity, but we also observe that the F1@$M$ metric is in conflict with the the goal of diversity to some extent. Especially for datasets with relatively fewer ground-truth keyphrases, results with higher diversity will decrease the F1@$M$ value.  Another line of work in KE task tends to uses $F_1$@5 and $F_1$@10 (without filling prediction) as evaluation metrics (see Appendix), $F_1$@5 is much higher than $F_1$@10 in most situations, which further proves that $F_1$@M is somewhat cannot reflect the diversity of results.

Further, measured by ranking metric, comparing under the very similar decoding setting with~\cite{DBLP:conf/acl/ChanCWK19} and the exactly same decoding setting with~\cite{DBLP:conf/sigir/SunTDDN19}, our model achieves an average of 5.5\% improvements against them.
This data also illustrates that our model can generate high-quality and diverse keyphrase lists.

\noindent \textbf{Ablation Study} We examine the performances of different model variants \textit{based on the same decoding settings} to evaluate the impact of different modules. 
From the results in Table~\ref{table:present-result}, we can see that: 
1) For both GCN and Div-GCN setting, dynamic computation mechanism brings stable performance improvement on $F_1$@$M$ (~2.5\% and ~6.9\% on KP20K and Krapivin) and significant improvement on $F_1$@5, which proves the effectiveness of the dynamic computation mechanism.
2) On most test sets, the diversified inference lifts the $F_1$@5 by a large margin but degrades the $F_1$@$M$ performances. It might because the DI process tends to generate longer sequences, which will cause the value of $F_1$@$M$ lower in some cases.
3) According to the rank-based metric NDCG@10, the DGCN encoder is essential to the model and achieves the best performances on three test sets. This result again proves that the inter-relation between keyphrases is important to the KE task.

\subsection{Discussion and Analysis}
\noindent \textbf{Dynamic Graph Properties}
To examine the effectiveness of the DGCN, we record some properties of the dynamic syntactic graphs during the inference process.
From the statistics in Table~\ref{table:graph-statistics}, we find that the graphs are very sparse since roughly only 2.6\% of the nodes-pairs have edges between them.
Besides, we calculate the average weights of the following three special edge types when decoding the $p$-th ($p\in [1, 2, \cdots, 7]$) keyphrase.
For edge $e_{ij}$ connected with node $i$ and node $j$: 
1) if both node $i$ and node $j$ are ground-truth target phrase words, we classify the edge as \textbf{E}$_{in\_in}$; 
2) if only one of the nodes lies in the ground-truth target phrase word set, this edge is called as \textbf{E}$_{in\_out}$;
3) all the other edges is classified to \textbf{E}$_{others}$.
As Figure~\ref{fig:graph_weight_trends} shows, the average weights of \textbf{E}$_{in\_in}$ and \textbf{E}$_{in\_out}$ types drops as the decoding process continues, while the average weights \textbf{E}$_{others}$ keeps steady. As Table~\ref{table:graph-statistics} shows, \textbf{E}$_{in\_in}$ accounts for no more than 10.8\% (On Inspec) of all the edges and E$_{in\_out}$ accounts for no more than 20.3\% (On Inspec). 
The statistics prove that the dynamic structure modifies the graph by reducing the edge weights of related nodes while keeping most of the edge weights unchanged.

\begin{table}[h]
\centering
\small
\begin{tabular}{l|cc|cc}
\hline \hline
\multicolumn{1}{c|}{\multirow{2}{*}{\textbf{Model}}} & \multicolumn{2}{c|}{\textbf{Inspec}} & \multicolumn{2}{c}{\textbf{SemEval}} \\
\multicolumn{1}{c|}{}                                & Avg. \#         & Corr. \#        & Avg. \#           & Corr. \#        \\
\hline \hline
oracle                                               & 7.20     &  -       &  6.12        &  -         \\
\hline
catSeqTG-$2RF_{1}$                                              & 3.45     &  1.41       &  3.73        &  1.48         \\
ExHiRD-h                                              & 4.00     &  N/A       &  3.65        &  N/A         \\
\hline
catSeqD                                              &    3.33  &   1.25      & 3.47     &    1.28       \\
GCN                                              & 3.53     &  1.55       &  3.77        &  1.45         \\
DGCN                                              &   3.58   &   1.60     &   4.05       &    1.58       \\
Div-DGCN                                              & \textbf{5.40}     &  \textbf{2.26}       &  \textbf{6.10}        &  \textbf{1.91}         \\
\hline
\end{tabular}
\caption{The abilities of predicting diversified and high-quality keyphrase set on two datasets. Avg. \# is the average number of generated keyphrases. Corr. \# is the average number of correctly predicted keyphrases.}
\label{table:number-of-keyphrases}
\end{table}

\noindent \textbf{Unique Keyphrase Numbers}
To better examine the diversity of generated keyphrase sequences, we study the average number of unique keyphrases as well as the corrected predicted keyphrase number to compare the predictive quality with several KE models.
Due to page limitation, we cannot list the results of KP20K in Table~\ref{table:number-of-keyphrases}, the average predicted keyphrase number and correct number in KP20K of our method is 4.90 and 1.45, much higher than which reported in ExhiRD-h~\cite{DBLP:journals/corr/abs-2004-08511} paper (3.97 and 0.81).
As Table~\ref{table:number-of-keyphrases} shows, compared to other models including our ablation variants, the Div-DGCN model generates much more keyphrases, which mitigates the insufficient generation problem, also the numbers of correctly predicted keyphrase are improved significantly (similar or higher precision when predicting more phrases).
Compared to our backbone model catSeqD, the DGCN model also generates longer and more diverse outputs.
The results illustrate the importance of the introduced syntactic-aware encoder and dynamic structure.

\noindent \textbf{Case Study}
We show a prediction case in Figure~\ref{fig:case_study}.
From the case, we observe that compared to the two SOTA models, our model predicts a more accurate keyphrase list.
Also, compared to the ablation variant, the predicted results with diversified inference cover more topics in the source document.
These results show that our model is captures the relation in the keyphrase set and achieves better results.

\section{Conclusion}
Modeling informative latent representations as well as capturing keyphrase dependencies is essential to Seq2Seq keyphrase extraction models. In this work we propose a Div-DGCN framework to address these two issues by enhancing the encoder with syntactic graph and modifying the graph edge information in the keyphrase decoding process. Further enhanced with a diversified inference process, the model can generate accurate and diverse outputs. Experimental results show that our model can surpass previous competitive models on various metrics and benchmark datasets, proving its effectiveness.

\clearpage
\bibliographystyle{aaai21}
\bibliography{emnlp2020}

\clearpage
\section*{Appendix A}

\begin{figure}[htbp]
\centering
\includegraphics[width=0.47\textwidth]{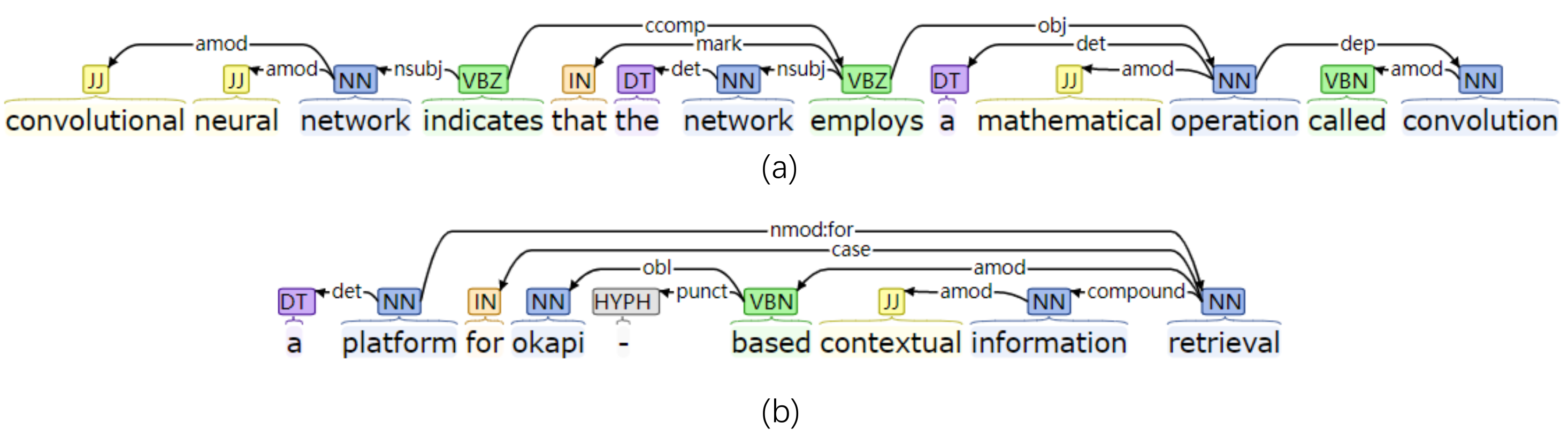}
\caption{Syntactic dependency tree for two samples.}
\label{fig:why_syntax_tree}
\end{figure}
\subsection*{Why using the syntactic information}
Syntactic tree helps to find possible candidate phrases in a sentence.
As Figure~\ref{fig:why_syntax_tree}-(a) shows, it captures "convolutional neural network" as a complete phrase rather than "neural network".
Moreover, as Figure~\ref{fig:why_syntax_tree}-(b) shows, the syntactic information provides a discontinuous phrase candidate "okapi based retrieval" based on long term dependencies.

\subsection*{Hyper-parameters}
\begin{table}[h]
\centering
\scalebox{1.0}{
\begin{tabular}{l l}
\hline
\textbf{Hyper-parameters} & \textbf{Value}  \\ \hline
BiGRU Layer & 1\\
Word Embedding Size ($d_w$) & 300 \\
Pos Embedding Size ($d_{pos}$) & 30\\
Position Embedding Size ($d_p$) & 10\\
GCN (DGCN) layer & 6 \\
GCN Embedding Size ($d_h$) & 400 \\
Edge Embedding Size ($d_e$) & 80\\
GRU Hidden Size & 400 \\
Decoder Layer & 3 \\
Phrase-Level Factor ($\lambda_1$) & 1.0\\
Beam-Level Factor ($\lambda_2$) & 0.1\\
\hline
\end{tabular}}
\caption{Hyper-parameters chosen for all models.}
\label{table:hyper}
\end{table}

\begin{table*}[t]
\centering
\scalebox{0.88}{
\begin{tabular}{l|p{0.8cm}p{0.8cm}| p{0.8cm}p{0.8cm} | p{0.8cm}p{0.8cm} | p{0.8cm}p{0.8cm} | p{0.8cm}p{0.8cm}}
\hline 
\hline
\multicolumn{1}{c|}{\multirow{2}{*}{\textbf{Model}}} & \multicolumn{2}{c|}{\textbf{Inspec}} & \multicolumn{2}{c|}{\textbf{Krapivin}} & \multicolumn{2}{c|}{\textbf{NUS}} & \multicolumn{2}{c|}{\textbf{SemEval}} & \multicolumn{2}{c}{\textbf{KP20k}} \\
\multicolumn{1}{c|}{} & $F_{1}@5$   & $F_{1}@10$   & $F_{1}@5$    & $F_{1}@10$    & $F_{1}@5$  & $F_{1}@10$ & $F_{1}@5$    & $F_{1}@10$   & $F_{1}@5$   & $F_{1}@10$  \\
\hline
CopyRNN & 0.278 & 0.342 & 0.311 & 0.266 & 0.334 & 0.326 & 0.293 & 0.304 & 0.333 & 0.262 \\
CatSeqD$(2018)$ & 0.276 & 0.333 & 0.325 & 0.285 & 0.374 & 0.366 & 0.327 & \textbf{0.352} & 0.348 & 0.298 \\
TG-Net$(2019)$    &0.315 & 0.381 & \textbf{0.349} & 0.295 & \textbf{0.406} & 0.370 & 0.318 & 0.322 & 0.372 & 0.315 \\
ParaNetT+CoAtt$(2019)$ & 0.296 & 0.357 & 0.329 & 0.282 & 0.360 & 0.350 & 0.311 & 0.312 & 0.360 & 0.289
\\
SKE-Large-Rank$(2020)$ & 0.300 & 0.334 & 0.313 & 0.264 & 0.400 & 0.379 & \textbf{0.356} & \textbf{0.351} & \textbf{0.392} & 0.328\\
\hline
Div-DGCN & \textbf{0.369} & \textbf{0.385} & \textbf{0.355} &\textbf{0.354} & \textbf{0.408} & \textbf{0.407} & 0.344 & 0.325 & 0.352 & \textbf{0.347} \\
\hline
\hline
\end{tabular}
} 
\caption{Results of present keyphrase prediction based on other metrics. SKE-Large-Rank uses the pre-trained BERT large.}
\label{table:another-results}
\end{table*}

\begin{table*}[t]
\centering
\scalebox{0.85}{
\begin{tabular}{l|cc| cc | cc | cc | cc}
\hline 
\multicolumn{1}{c|}{\multirow{2}{*}{\textbf{Model}}} & \multicolumn{2}{c|}{\textbf{Inspec}} & \multicolumn{2}{c|}{\textbf{Krapivin}} & \multicolumn{2}{c|}{\textbf{NUS}} & \multicolumn{2}{c|}{\textbf{SemEval}} & \multicolumn{2}{c}{\textbf{KP20k}} \\
\multicolumn{1}{c|}{} & $F_{1}@M$   & $F_{1}@5$   & $F_{1}@M$    & $F_{1}@5$    & $F_{1}@M$  & $F_{1}@5$ & $F_{1}@M$    & $F_{1}@5$   & $F_{1}@M$   & $F_{1}@5$  \\
\hline
GCN   & 0.312$_{7\ }$ & 0.261$_{4\ }$ & 0.331$_{9\ }$ & 0.262$_{5\ }$ & 0.391$_{13}$ & 0.325$_{14}$& 0.303$_{9\ }$ & 0.264$_{9\ }$& 0.350$_{4\ }$ & 0.281$_{2\ }$ \\
Div-GCN   & 0.360$_{8\ }$ & 0.331$_{6\ }$ & 0.324$_{5\ }$ & 0.280$_{3\ }$ & 0.389$_{7\ }$ & 0.371$_{11}$& 0.317$_{6\ }$ & 0.289$_{4\ }$& 0.343$_{1\ }$ & 0.304$_{2\ }$ \\
DGCN   &0.327$_{7\ }$ &0.275$_{8\ }$ & 0.354$_{7\ }$ & 0.284$_{4\ }$ & 0.397$_{9\ }$ & 0.328$_{2\ }$& 0.319$_8$ & 0.280$_{12}$ & 0.359$_0$ & 0.290$_{4\ }$ \\
Div-DGCN   &\textbf{0.376$_{3\ }$} &\textbf{0.348$_{6\ }$} & 0.345$_{2\ }$ & \textbf{0.313$_{4\ }$} & 0.391$_{7\ }$ & \textbf{0.379$_{6\ }$}& 0.323$_{9\ }$ & \textbf{0.320$_{14}$}& 0.349$_{2\ }$ & \textbf{0.313$_{3\ }$} \\
\hline
\hline
\multicolumn{11}{c}{\textbf{Normalized Discounted Cumulative Gain @10 (NDCG@10)}}\\
\hline
GCN & \multicolumn{2}{c|}{0.657$_{5\ }$} & \multicolumn{2}{|c|}{0.582$_{2\ }$} & \multicolumn{2}{|c|}{0.787$_{9\ }$} & \multicolumn{2}{|c|}{0.677$_{15}$} & \multicolumn{2}{|c}{0.590$_{6\ }$}\\
\hline
DGCN & \multicolumn{2}{c|}{0.663$_{3\ }$} & \multicolumn{2}{|c|}{0.597$_{3\ }$} & \multicolumn{2}{|c|}{\textbf{0.797$_{7\ }$}} & \multicolumn{2}{|c|}{\textbf{0.704$_{10}$}} & \multicolumn{2}{|c}{\textbf{0.616$_{5\ }$}}\\
\hline
Div-DGCN & \multicolumn{2}{c|}{\textbf{0.692$_{4\ }$}} & \multicolumn{2}{|c|}{\textbf{0.611$_{8\ }$}} & \multicolumn{2}{|c|}{\textbf{0.797$_{14}$}} & \multicolumn{2}{|c|}{0.701$_{16}$} & \multicolumn{2}{|c}{\textbf{0.620$_{1\ }$}}\\
\hline
\end{tabular}
} 
\caption{
Results with standard deviation on all datasets.
}
\label{table:results_variance}
\end{table*}

\subsection*{Data sources}
We use the predictions released publicly\footnote{https://github.com/kenchan0226/keyphrase-generation-rl} to calculate CatSeqTG-2RF$_1$'s NDCG@10 and average predicted/correct keyphrase numbers. We use the average predicted keyphrase numbers of ExHiRD-h as reported by its original paper.

\subsection*{Corpus and Evaluation}
We use the public available pre-processed datasets~\footnote{https://drive.google.com/file/d/1DbXV1mZXm\_o9bgfwPV9PV0\\ZPcNo1cnLp/view}.
Also, we follow the previous work's evaluation process~\footnote{https://github.com/kenchan0226/keyphrase-generation-rl/blob/master/evaluate\_prediction.py}.

\subsection*{The Diversified Inference Process}
Due to the page limitation, we put the commented DI process here to show the procedure more clearly.

\begin{algorithm}[h]
\caption{Inference Procedure}
\label{algo:1-comment}
\small
 Input:\\
 - $\mathcal{D} = \{x_1, \cdots, x_l\}$ ; $\textbf{y}^{<p} = \{y^1, \cdots, y^{p-1}\}$ \\
 - Model parameters: $\theta$ \\
 
 Perform beam search for the $p$-th keyphrase using beam width $B$ \\
 $H^p, c^p = \mbox{ENCODER}(D, \textbf{y}^{<p}; \theta)$ \\
\For{$t=1,\ \ldots \,T$}{
    \tcp{predictive probability for $B$ groups over vocabulary $\mathcal{Y}_d$}
    $P(y^p_{b, t}) = \mbox{DECODER}(y^p_{b, <t}, H^p, c^p; \theta)$ \\
    \tcp{$B\cdot |\mathcal{V}_d|$ candidate sequences}
    $\mathcal{Y}^p_{t} =  \mathcal{Y}^p_{t-1} \times P(y^p_{b, t})$ \\
    \tcp{calculate diversified sequence scores}
    $\hat \Theta(\textbf{y}_{b,[t]}^p) \leftarrow \Theta(\textbf{y}_{b,[t]}^p) + \lambda_1 \mbox{DP}(\textbf{y}_{b,[t]}^p, \textbf{y}^{<p}) - \lambda_2 \mbox{SP}(y^p_{b, t})$  \\ 
    \tcp{perform one step of beam search}
        $Y_{t}^p \leftarrow \mathop{\arg\max}_{(\textbf{y}_{1,[t]}^p, \dots, \textbf{y}_{B,[t]}^p)} \sum_{b\in[B]}\hat \Theta(\textbf{y}_{b,[t]}^p)$ \\
}
Return $y^p = \mathop{\arg\max} (Y^p_{T})$ using $\hat \Theta(\cdot)$ as the predicted $p$-th keyphrase.
\end{algorithm}

\subsection*{Repeated Experiments}
We conduct the experiment on each model variant for 3 times using different random seeds, and report the average performances in the paper for brevity.
Table~\ref{table:results_variance} lists the average and deviation values of each model on several metrics. The subscript represents the corresponding standard deviation value (e.g., $0.342_3$ indicates $0.342\pm 0.003$).

\subsection*{Impact of the Diversified Inference Factor}
\begin{figure}[h]
\centering
\includegraphics[width=0.45\textwidth]{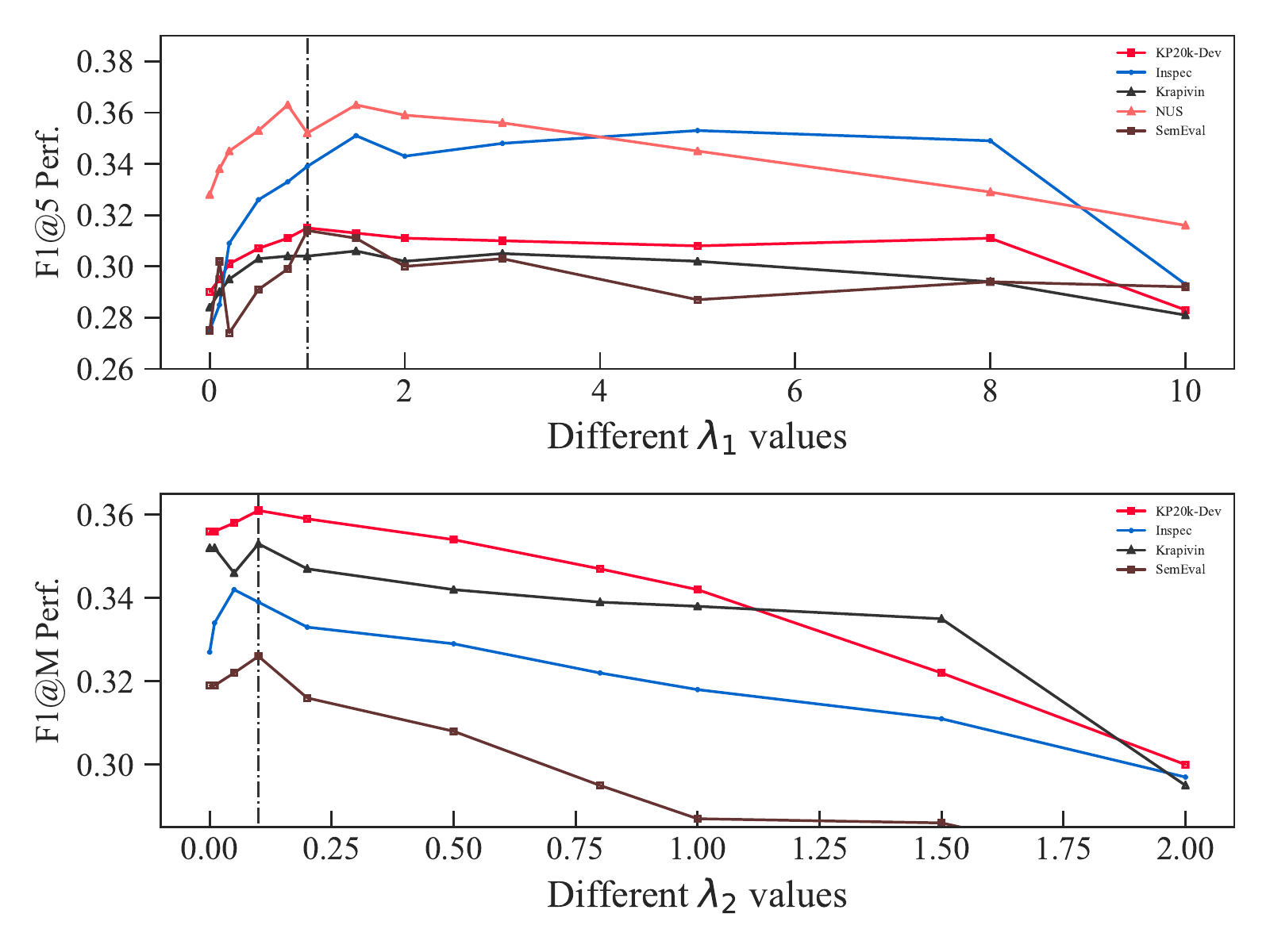}
\caption{Performance curves of phrase-level factor $\lambda_1$ and beam-level factor $\lambda_2$.}
\label{fig:perf_curve}
\end{figure}

We search the two factors on Kp20k dev set seperately, and use the combination of the two best factors. 
Our search space for the phrase-level factor $\lambda_1$ is $[0.1, 0.2, 0.5, 0.8, 1.0, 1.5, 2.0, 3.0, 5.0, 8.0, 10.0]$ and the search space for the beam-level factor $\lambda_2$ is $[0.01, 0.05, 0.1, 0.2, 0.5, 0.8, 1.0, 1.5, 2.0]$.
The performance curves on KP20k dev set and several test sets are shown in Figure~\ref{fig:perf_curve}.
We observe that $\lambda_1$ improves the $F_1$@5 by a large margin but degrades $F_1$@M, and $\lambda_2$ boosts the $F_1$@M performances, so we choose these two metrics in Figure~\ref{fig:perf_curve}.

\subsection*{Comparison based on Other Metrics}
In keyphrase extraction task, some studies choose to use the $F_1@M$ and filled version of $F_1@5$ as evaluation metrics (we choose these two metrics) while others evaluate the performances using $F_1@5$ and $F_1@10$. The latter $F_1@5$ does not fill random wrong predictions when prediction number is less than 5.
To thoroughly evaluate the effectiveness of our model, we further compare the full model's performance on $F_1@5$ and $F_1@10$ in Table~\ref{table:another-results}.
Baseline methods listed in Table~\ref{table:another-results} only report these two metrics so we cannot compare to them in the experiments section of our paper.
We find that our model exceeds the baselines by a relative large margin especially in $F_1@10$.
Note that SKE-Large-Rank surpasses our method on several metrics but it uses the pretrained language model BERT-LARGE and thus takes the advantages on document representations.

\end{document}